\documentclass[runningheads]{llncs}

\usepackage{eccv}

\usepackage{eccvabbrv}

\usepackage{graphicx}
\usepackage{booktabs}
\usepackage{tcolorbox}
\tcbuselibrary{skins}
\usepackage{listings}

\usepackage[accsupp]{axessibility}  % Improves PDF readability for those with disabilities.

\usepackage[pagebackref,breaklinks,colorlinks,citecolor=eccvblue]{hyperref}
\usepackage{enumitem}
\usepackage{orcidlink}
\usepackage{multirow}
\usepackage{wrapfig}

\newcommand{\method}{\textsc{PQSG}}
\newcommand{\dataset}{\textsc{FinePhyEval}}

\usepackage[capitalize]{cleveref}
\crefname{section}{Sec.}{Secs.}
\Crefname{section}{Sec.}{Secs.}
\Crefname{table}{Tab.}{Tabs.}
\crefname{table}{Tab.}{Tabs.}
\Crefname{figure}{Fig.}{Figs.}
\crefname{figure}{Fig.}{Figs.}
\crefname{appendix}{Sec.}{Secs.}
\Crefname{appendix}{Tab.}{Tabs.}

\usepackage[framemethod=TikZ]{mdframed}
\definecolor{OurColor}{HTML}{36aa70}
\definecolor{UserExampleBg}{HTML}{ffffff}
\definecolor{UserExampleTitle}{HTML}{545f7f}
\newmdenv[
    roundcorner=5pt,
    backgroundcolor=UserExampleBg,
    linecolor=UserExampleTitle,
    outerlinewidth=0.5pt,
    frametitlebackgroundcolor=UserExampleTitle,
    frametitlefont={\bfseries\color{white}},
]{user_example}

\begin{document}

\title{Physics Question Scene Graph: Fine-grained Evaluation of Physical Plausibility in Text-to-Video Generation} 

\titlerunning{Physics Question Scene Graph (PQSG)}

\author{Atin Pothiraj\inst{1} \and
Jaemin Cho\inst{2,3}\orcidlink{0000-0002-1558-6169} \and
Yue Zhang\inst{1}\orcidlink{0000-0003-2153-6536} \and \\
Elias Stengel-Eskin \inst{4}\orcidlink{0000-0002-6689-505X} \and
Mohit Bansal \inst{1}\orcidlink{0009-0009-2965-5354}
}

\authorrunning{A. Pothiraj et al.}

\institute{University of North Carolina at Chapel Hill, USA  \and
 AI2, USA \and Johns Hopkins University, USA \and
 University of Texas at Austin, USA \\
\url{https://github.com/atinpothiraj/pqsg}
}

\maketitle

\begin{abstract}
Video generation models are increasingly capable of producing realistic videos, but they still struggle to generate videos that follow basic physical laws.
Compounding this is a lack of reliable granular evaluation methods for localizing and specifying physical law violations in videos. 
We address this by introducing Physics Question Scene Graph (\method{}), a hierarchical question-based evaluation pipeline. 
\method{} evaluates generated videos by checking their faithfulness to a prompt across objects, actions, and adherence to physical laws using a graph-based hierarchy of questions generated by a vision-language model (VLM), guided by high-quality in-context examples.  
By representing questions as a graph, \method{} introduces logical dependencies within questions, ensuring that each query is contextually valid. 
Moreover, \method{} provides granular assessments of which qualities of the video violate physical plausibility constraints. 
We validate \method{} by creating \dataset{}, a dataset with physics-based prompts and corresponding generated videos from diverse state-of-the-art video generation models (Sora 2, Veo 3, and Wan 2.1), with each video annotated across multiple categories by humans. 
Using \dataset{}, we measure the correlation between \method{}'s fine-grained scores and human judgments, showing higher overall correlations than prior work.
We also find that \method{} ranks closed-source models higher than Wan 2.1 on physical realism.
Lastly, we show that the annotations we provide in \dataset{} can also be used for subtask evaluation: we benchmark two strong VLMs on generating and answering questions, finding that while models can create human-like questions, they still fall short of human performance in answering them. 
\keywords{Physics \and Evaluation \and Video \and Text-to-Video Generation}

\end{abstract}
\section{Introduction}

\begin{figure*}
    \centering
\includegraphics[width=\linewidth]{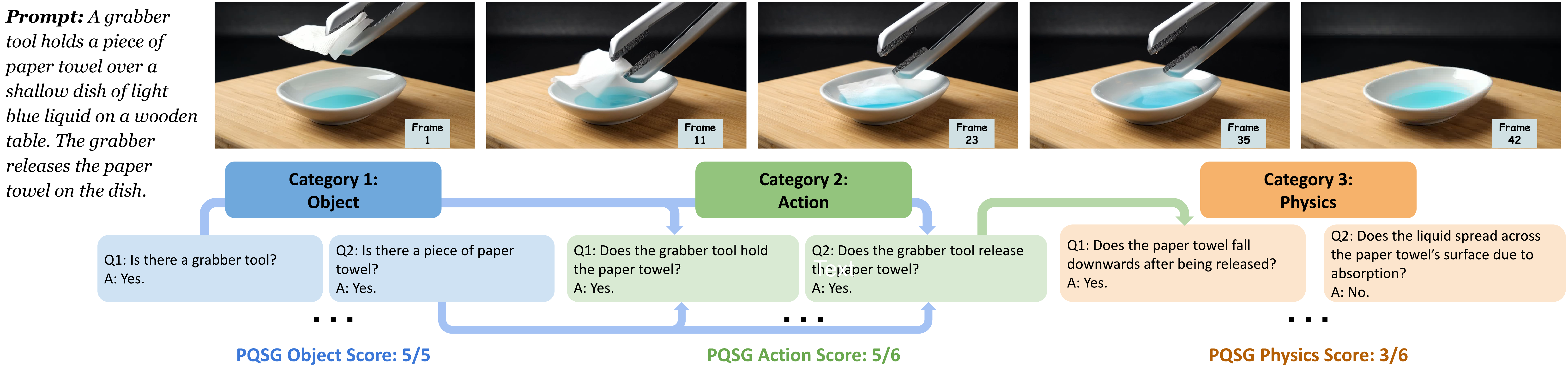}
    \caption{
    An example prompt and its corresponding generated video from Wan 2.1, with sample \method{} nodes and edges (not all are included)  for each category.
    While the video contains the correct objects (a grabber, paper towel, bowl, etc.) it does not show all the right actions, and the physical interactions shown are implausible, with the paper towel dissolving into the liquid rather than absorbing it.
    \method{} specifies in which category the video is unrealistic (in this case, in its physics) and, within that category, which physical interactions are implausible (here, that the liquid fails to absorb into the paper towel, reflected in Physics Q2).
    }
    \label{fig:teaser}
\end{figure*}

An intuitive understanding of physical laws has been posited as a core component of human cognition \cite{lake2017building}, allowing us to reason accurately about the world and the consequences of our actions in it. 
This capacity, often called ``world modeling,'' has long been a goal of video generation models \cite{sora_2024, qinworldsimbench}: to reason about future world states by rendering future video frames. 
Indeed, recent video generation models have become increasingly adept at generating photorealistic videos of objects and people in various environments \cite{blattmann2023stable}.
However, while some understanding of basic physical laws has been documented even in infants \cite{spelke1990principles, spelke1995development}, current video generation models still struggle with physical realism, producing outputs that violate fundamental physical laws, such as solid mechanics, fluid dynamics, and optics \cite{motamed2025generative}. 
Adherence to these laws is key for developing true world models that can serve as practical simulators.
For example, a simulation of dish-washing must accurately represent water flow and the contact between a sponge and a dish.
Such physical accuracy is critical for downstream applications in robotics \cite{fu2025learning}, embodied agent training \cite{soni2025videoagent}, and synthetic data generation \cite{choi2025svad}.

Compounding current models' shortcomings in producing physically-plausible videos is a lack of reliable, fine-grained, and automated methods for evaluating physics adherence.
Existing methods provide only high-level, coarse-grained scores, failing to capture the inherent structure of physics evaluation. 
We argue that a robust physics evaluation must capture the logical dependencies within a scene.
For example, to evaluate whether an object's  \textit{fall} is physically plausible, one must verify that the object is \textit{present} (object-level correctness) and that it is \textit{falling} (action-level correctness). Only then can the \textit{physics} of the fall (e.g., acceleration) be assessed.
Existing methods (e.g., \cite{motamed2025generative, fid2017, huang2024vbench}) lump these factors into a single, aggregate score, making it difficult to determine \emph{why} a video failed. 
Moreover, these combined metrics are often fooled by videos that are visually realistic and temporally consistent, but may still be physically implausible \cite{yin2026survey}. 
For example, in \cref{fig:teaser},
baseline methods may assign a high score to a video where human annotators identify a clear physics violation (e.g., a paper towel dissolving instead of absorbing liquid).
As models become more powerful, the errors they display will also become subtler, motivating the need for a fine-grained evaluation metric that can provide feedback on specific physical inaccuracies, both for evaluation and for fine-grained video repair \cite{lee2024self}.  

To address this gap, we introduce the \textbf{P}hysics \textbf{Q}uestion \textbf{S}cene \textbf{G}raph (\method{}), a novel, structured evaluation framework for video generation models that measures fine-grained details of physical dynamics. 
\method{} is designed with two key properties: 
(1) \method{} is \textbf{granular}: rather than assessing adherence to physical laws in one, overarching score, it decomposes the assessment into multiple sub-queries, allowing for granular feedback on exactly which aspect of the video violates a physical law. 
(2)  \method{} follows a \textbf{dependency structure}, ensuring that questions are asked in a sensible way: if an object is not present in a video, we do not subsequently ask about actions or physics interactions involving it. 
We illustrate this in \cref{fig:teaser}, where the sub-queries are divided into three hierarchical categories: \textbf{object}, \textbf{action}, and \textbf{physics}. 
Note that PQSG evaluates the semantics \textit{mentioned in text prompts} --- it assesses whether a generated video realizes the objects, actions, and physical interactions specified by its prompt.
We formulate \method{} as a two-stage process: first, in the question-generation stage (QG), we task vision-language models (VLMs) with generating hierarchical physics-aware graphs of questions from text prompts. 
In the second stage -- question-answering (QA) --  these physics scene graphs are then passed with the generated video to a VLM to answer questions about the video, producing interpretable object, action, and physics scores, as seen in \cref{fig:teaser} (as well as \cref{fig:video_eval_examples}).
In the case of \cref{fig:teaser}, \method{}'s QG stage generates targeted questions (e.g., questions about fluid dynamics) which are then answered, leading to appropriately low physics scores. 

To evaluate the components of \method{} (QG, QA) and subsequently evaluate diverse video generation models using \method{}, we collect \dataset{}, a dataset of 195 human-annotated prompt-video pairs, with prompts sourced from Physics-IQ \cite{motamed2025generative} and videos generated from three diverse state-of-the-art video generation models: Sora 2 \cite{openai2025sora2}, Veo 3 \cite{google2025veo}, and Wan 2.1 \cite{wan2025}.
\dataset{} provides comprehensive human annotations for: (1) video quality across four Likert-scale categories (object, action, physics, and overall quality), where we find strong inter-annotator agreement; and (2) \method{}'s constituent subtasks of question-generation (QG) and question-answering (QA).

On \dataset{}, we first show that \method{} results in better correlations to overall human ratings than prior metrics, and that -- unlike prior metrics -- \method{} can provide fine-grained evaluation feedback at the level of individual attribute categories.
Moreover, we rank different models according to \method{}'s automated metric, finding that proprietary models (Sora 2, Veo 3) outperform the open-source models (Wan 2.1, Cosmos-14B), and that all models struggle on action and physics more than object generation. 
We also ablate the components of \method{}, measuring the contribution of fine-grained questions and dependency graphs. 
Additionally, the annotations we provide in \dataset{} allow for low-level subtask evaluation on QA and QG: we benchmark two strong VLMs -- Gemini-2.5-Pro \cite{gemini25} and GPT-5.5 \cite{openai2025gpt5} -- on their ability to perform QG and QA, finding models to be adequate at QG but well short of human performance on QA despite the relatively short length of videos (avg. length: 4.39 seconds).
In the overall evaluation, we find that for QA, action and physics categories pose challenges to models, with the best model, GPT-5.5, obtaining $\sim65\%$ QA accuracy.
This suggests promising directions for improvement in evaluation: 
we show that \method{} has a high upper bound correlation with human judgments when QA is performed by a human, suggesting further gains will be obtained as VLMs improve. 

\section{Related Work}

\noindent\textbf{Video Generation Models.}
Progress in text-to-video generation has accelerated rapidly with the advent of video diffusion models~\cite{hong2022cogvideo, ho2022imagen, ho2020denoising, yangcogvideox, bao2024vidu, ni2024ti2v, wang2026anchorweave, wang2025epic}.
Diffusion models can now realistically render complex high-resolution scenes, with recent examples approaching the realism of human-shot videos~\cite{google2025veo, openai2025sora2}. 
However, recent studies indicate that even state-of-the-art models frequently violate basic physical principles~\cite{bansal2025videophy,motamed2025generative}. 
To address this, previous research has proposed various physics-aware enhancements, including motion guided by the physics-simulator~\cite{liu2024physgen, tan2024physmotion, montanaro2024motioncraft}, physics-informed post-training and reward optimization~\cite{li2025pisa, wang2025physcorr, lin2025reasoning, huang2026phymotion}, implicit physical priors injected through vision-language reasoning or force-based conditioning~\cite{yang2025vlipp, wang2025physctrl, gillman2025force, hao2025enhancing, huang2025planning}, and program-based generative models that explicitly encode physical laws~\cite{liu2024physgen}. 
Critically, making progress towards more physically-realistic models requires a fine-grained evaluation framework that offers precise, actionable metrics on \textit{which} physical laws and phenomena current models struggle with.  

\noindent\textbf{Evaluation of Generated Videos.}
Early evaluation metrics for image and video generation have focused primarily on general visual quality or semantic alignment with textual prompts, employing methods based on deep feature embeddings~\cite{fid2017, unterthiner2019fvd}, classifier predictions~\cite{salimans2016improvedtechniquestraininggans}, semantic matching~\cite{hessel2021clipscore}, or comprehensive evaluation toolkits~\cite{liu2023evalcrafter}. However, these metrics do not explicitly measure physical realism or pinpoint precise violations of physical laws, as their criteria inherently overlook fine-grained physical constraints and nuanced action semantics.
Recent evaluation frameworks for video generation have attempted to address physical realism, prompt alignment, and factual consistency through several approaches. Methods such as VideoScore~\cite{he2024videoscore} primarily evaluate semantic and factual alignment, but inherently overlook subtle yet critical violations of physical laws. More specialized methods, such as VideoPhy-2~\cite{bansal2025videophy}, try to mitigate this gap by fine-tuning vision-language models (VLMs) to produce semantically meaningful realism scores. 
In a similar vein, PhyGenEval~\cite{meng2024towards} leverages VLMs for assessing physical commonsense.
Our work differs from prior efforts in its focus on structured, fine-grained evaluation.
Unlike prior work that provides a single aggregate score or a few coarse-grained scores, \method{} generates a hierarchical set of detailed questions that localize exactly where a generated video should be improved.
This granular QA format makes \method{} easily interpretable and ensures that evaluations are consistent across answerers, human-aligned, and generalizable.
In a similar spirit to our work, DSG~\cite{cho2024davidsonian} offers question-based evaluation using graphs. 
However, DSG is designed for images and fails to test for temporal properties (such as actions or physics-based interaction), as shown in our evaluation (see \cref{sec:exp_correlations}).

\section{Physics Question Scene Graph (\method{})}

Given a text prompt and a generated video, we aim to output video scores across multiple dimensions with a special focus on adherence to physical laws and with the goal of developing an evaluation system that identifies the specific failure points in generated videos, enabling fine-grained analysis.
For example, consider the prompt: \textit{``Two pillows on a table and two grabber tools hanging above them from which a brown tennis ball and an orange block are suspended. The grabber tools let go of the ball and block.''} 
One key physics-related moment is the tennis ball falling, and we want to examine whether the fall seems plausible in accordance with the laws of gravity.
However, evaluating this ``plausible fall'' presents two significant challenges.
On one hand, many general video evaluation frameworks focus on foundational elements like object existence (e.g., ``Is there a ball?'') and attribute binding (e.g., ``Is the ball brown?'').
These methods are often insufficient for evaluating complex object dynamics -- the core of a physics evaluation.
On the other hand, for a framework to successfully judge the dynamics of the fall, it must first verify its preconditions: (a) that the tennis ball exists, and (b) that the ball is being dropped.
If either of these preconditions fails, the video generation model has failed before even attempting the physics, rendering a judgment on the ``fall's realism'' impossible and making such judgments potentially hallucinated or misleading.
Previous frameworks have generally failed to model this specialization in dynamics and these prerequisite dependencies. 
To jointly address both challenges, we introduce \method{}.

\begin{figure}[t]
    \centering
    \includegraphics[width=0.88\linewidth]{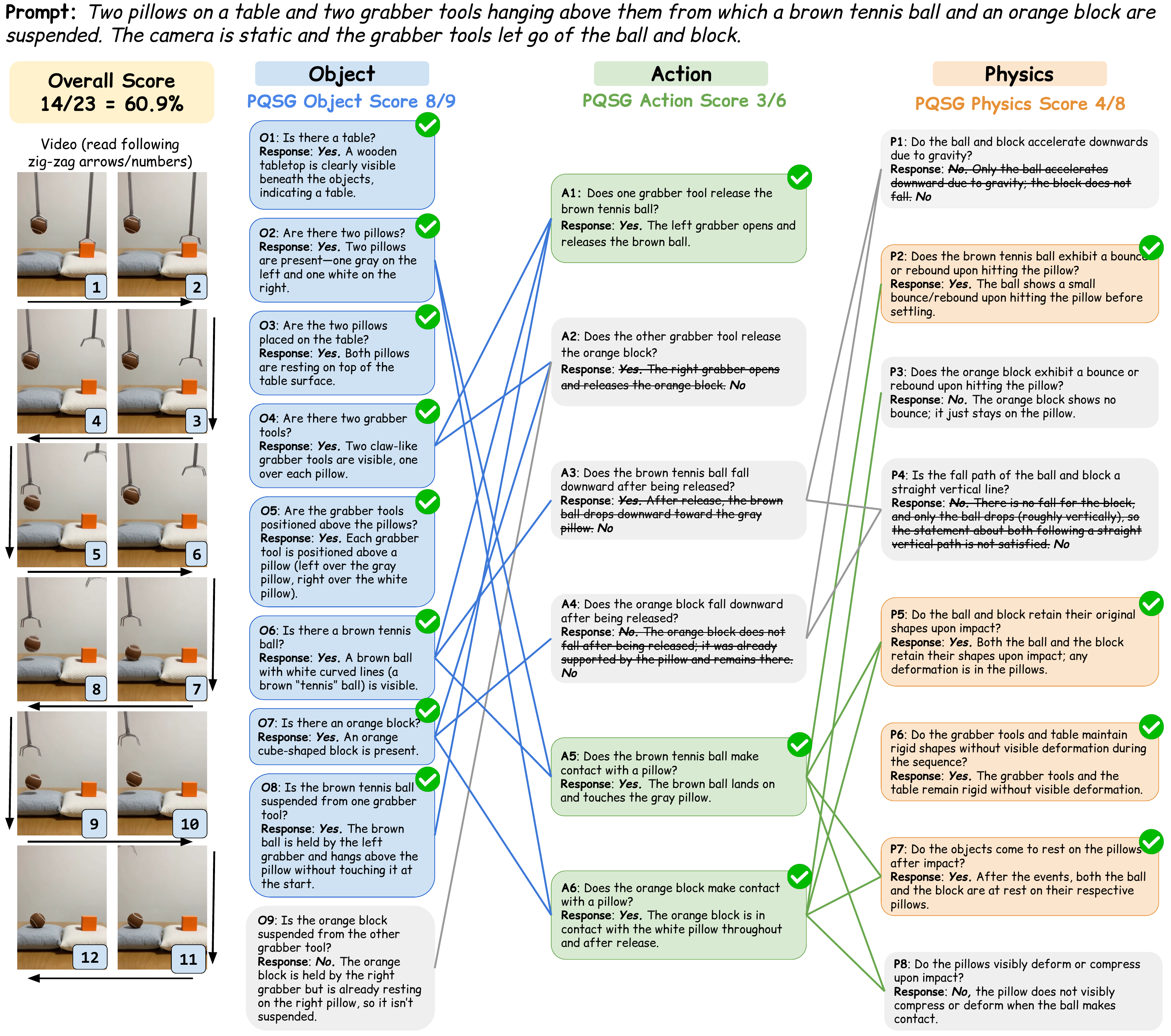}
    \caption{Illustration of fine-grained evaluation on a generated video with \method{}.
    For each question (e.g., 
    P8: \emph{``Do the pillows visibly deform or compress upon impact?''}), 
    \method{} provides binary judgment and reasoning behind each judgment (\emph{``A: No, the pillow does not visibly compress or deform when the ball makes contact''}).
    When a parent question is answered \emph{``no,''} then its children's questions are automatically also marked as \emph{``no''}
    (e.g., node A2 becomes invalid because its parent node, O9, is answered with \emph{``no''}).
    From the per-question evaluation result, we obtain per-category scores and an overall score, which is the sum of the per-question scores.
    The fine-grained questions pinpoint the detailed failure modes of generated videos.
    }
    \label{fig:video_eval_examples}
\end{figure}

The \textbf{Physics Question Scene Graph (\method{})} is a framework for fine-grained evaluation of a generated video by defining the physical scene as a directed acyclic graph (DAG).
Following DSG~\cite{cho2024davidsonian}, this graph consists of atomic verification questions (nodes) with explicit logical dependency structures (edges), all generated from the text prompt describing the scene.
We show an example video evaluation with \method{} in \cref{fig:video_eval_examples}.

\noindent\textbf{Nodes.}
\method{} organizes nodes into three hierarchical categories that can produce separate, fine-grained scores:
\begin{itemize}
\item \textbf{Object}: Verifies that a key object from the prompt is present in the video (e.g., \emph{``O2: Are there two pillows?''}).
\item \textbf{Action}: Verifies the object is exhibiting the correct action (e.g., \emph{``A1: Does one grabber tool release the brown tennis ball?''}).
\item \textbf{Physics}: Evaluates the plausibility of actions regarding physical laws (e.g., \emph{``P8: Do the pillows visibly deform or compress upon impact?''}).
In general, actions and physics are distinguished by whether they are mentioned in the prompt: actions are more directly tied to the prompt, whereas physics interactions (e.g., deformation, absorption, etc.) are implicit and part of physical commonsense.  \end{itemize}
This explicit separation of \textbf{Action} node from \textbf{Physics} node is what allows our method to pinpoint failures by 
isolating physical plausibility from simpler generation failures:
(a) a video fails to show an action in the first place \textit{vs.} (e.g., an object does not fall)
(b) a video shows an action, but the action may not look plausible regarding physical laws (e.g., an unrealistic fall).

\noindent\textbf{Edges.}
\method{} organizes questions into a hierarchical graph encoding their dependencies. 
As illustrated in \cref{fig:video_eval_examples}, most of the action binding nodes have object existence nodes as parents, and most of the physics nodes have action binding nodes as parents.
This hierarchy ensures that an object must exist and undergo the correct action before its physics can be judged.
Beyond these cross-level dependencies, PQSG also admits same-category edges like Object-to-Object, Action-to-Action, and Physics-to-Physics, which capture sequential dependencies within a category, such as a later action that presupposes an earlier one, or a physical outcome that is contingent on a preceding physical state. Edges are thus drawn from object to action, from action to physics, or between nodes of the same category, and no other connections are permitted. 
During evaluation, this dependency structure is strictly enforced. If the answer to a parent node's question is ``no'' (e.g., the Object node fails), the questions from its entire chain of child nodes are not queried and are automatically also marked with ``no.''
A child question is only queried if all its prerequisite parent nodes have been answered affirmatively. 
This ensures that only answerable questions are actively posed to the model, reducing the potential for hallucination. 
As shown in our ablation study (\Cref{tab:ablation}), this dependency structure yields stronger human correlation than unstructured approaches. 

\noindent\textbf{Evaluation Pipeline.}
Our evaluation pipeline consists of two steps: Question Generation (QG) and Question Answering (QA).
Both naturally lend themselves to VLM implementations (e.g., via Gemini 2.5 Pro~\cite{gemini25}, GPT-5.5~\cite{openai2025gpt5}).

In the QG step, we prompt the VLM with task instruction and one in-context example of a prompt, \method{} nodes, and \method{} edges, followed by a new prompt we want to generate \method{} from. See \cref{fig:qg_prompt} for the QG prompt.
In the QA step, we prompt a VLM with a generated video and a \method{} node (a question), and let the model answer one question at a time.
In our initial experiments, we find that letting the QA model answer only with yes/no responses often discourage the use of chain-of-thought reasoning. Instead, we implement QA in two steps: (a) generating an open-ended response and (b) categorizing the answer as ``yes'' or ``no,'' using the same VLM for both steps. 
We find this strategy improves the QA quality while adding negligible latency.
After marking responses from invalid nodes as ``no,'' we take the ratio of the ``yes'' response as the average score. In addition to the average score, we can also calculate separate, fine-grained scores for each of the three node categories (Object, Action, and Physics).

In \cref{fig:video_eval_examples}
we illustrate an evaluation of a generated video with \method{}.
\method{} provides per-question level fine-grained evaluation in binary judgment and reasoning behind each judgment.
From the per-question evaluation result, we can obtain per-category scores and an overall score, which is the sum of all per-question scores.
Furthermore, the fine-grained questions help pinpoint the detailed failure modes of generated videos.
\section{FinePhyEval: Human Annotation of Fine-grained Video Scores}
\label{sec:dataset}

To analyze the utility of fine-grained video evaluation by \method{}, we collect \dataset{}, a new dataset of human judgments across fine-grained categories for videos generated by recent SoTA video generation models.

\subsection{Prompt and Video Generation}
\noindent\textbf{Evaluation Prompts \& Reference Videos.}
We source our prompts from Physics-IQ~\cite{motamed2025generative} because it contains prompts designed to test video generation models' understanding of physical principles and also provides a reference ground-truth video for each prompt.
We utilize all 65 prompts from the dataset; an example prompt is shown in \cref{fig:video_eval_examples}, highlighting the prompts' complexity and compositional nature, with multiple objects and object interactions.

\noindent\textbf{Video Generation Models.}
While the original Physics-IQ paper reports the performance of older video generation models (e.g., Sora, Runway, Stable Video Diffusion), recent models demonstrate markedly improved capabilities.
To validate our evaluation system with the most recent and most visually realistic models, we collect videos from three recent state-of-the-art models, Sora 2~\cite{openai2025sora2}, Veo 3~\cite{ google2025veo}, and Wan 2.1~\cite{wan2025}, based on their high rankings on the Artificial Analysis text-to-video leaderboard~\cite{aa_arena2025} (as of March 2026) and their public API/checkpoint availability.
We additionally evaluate Cosmos-Predict2.5-14B~\cite{ali2025world}, a model purpose-built as a physical world simulator, to test whether an architecture designed for physical realism yields more physically plausible videos under our fine-grained evaluation.
We generate 260 prompt-video pairs (65 per model) using their default configurations.
This yields 4-second, 720x1080 (30 fps) videos from Sora 2; 4-second, 1280x720 (24 fps) videos from Veo 3; 5-second, 1280x720 (16 fps) videos from Wan 2.1; and 6-second, 1280$\times$704 (16 fps) videos from Cosmos-Predict2.5-14B.

\subsection{Human Annotation of Question Generation and Answering}
\noindent\textbf{Human Annotation of Fine-grained Verification Questions.}
To establish a ground-truth set of verification questions -- i.e., to establish ground-truth question generation (QG) -- we manually annotate verification questions for a sample of 20 prompts from \dataset{}. 
Following DSG~\cite{cho2024davidsonian},
we ensure that the questions cover the full content from the prompt, while each question is atomic (asking only one aspect of the prompt at a time) and does not overlap with another question.
We later measure the precision and recall of automatically-generated questions from the QG system against these questions. 

\noindent\textbf{Human Annotation of Answers to Verification Questions.}
To establish an upper-bound for question-answering (QA) performance, we collect human yes/no answers to the generated questions on 30 prompt-video pairs sampled for the QG verification step. 
In total, we manually annotate 444 QA pairs, as each prompt-video pair has a \method{} with multiple questions in it.

\subsection{Likert-scale Video Judgments}
\noindent\textbf{Human Annotation of Likert-scale Video Scores across Four Categories.}
To validate the overall scores produced by different methods, we collect human Likert score judgments over generated videos. 
Specifically, for each of 195 videos in \dataset{}, we collect four text-video alignment scores (object, action, physics, and overall categories), with eight non-author human annotators; i.e., 4 $\times$ 195 = 780 scores in total. 
Each score is measured in Likert-scale (1-5).
Object, action, and physics categories measure object existence, high-level action faithfulness, and physical plausibility, respectively.
The overall category measures an annotator's judgment of video-text alignment without focusing on the three criteria.
We study how this overall judgment Likert score is correlated with
different video evaluation metrics (\Cref{tab:correlation_human_method_comparison})
and 
category-specific human Likert scores (\Cref{tab:correlation_human_only_category}).
Note that this four-category human judgment for text-video alignment is more fine-grained than many previous works that asks annotators to consider one or two categories;
we study correlation between per-category human Likert score and per-category \method{} score in \Cref{tab:correlation_human_metric_category}.
We also calculate the intraclass correlation coefficient (ICC) on a subset of \dataset{} and find a high inter-annotator agreement of 0.84 across categories (see \cref{tab:agreement}).
See the \cref{sec:annotation_details} for annotation details, including full annotator guidelines.

\section{Experiments and Discussion}
\label{sec:experiments}
We first compare our \method{} with other single summary score methods in terms of human judgments
(\cref{sec:exp_correlations}).
Then we show the comparison of video generation models in \method{} scores (\cref{sec:exp_comparing_video_models}).
We also analyze the reliability of two subtasks of \method{}:  question generation and question answering  (\cref{sec:exp_subtasks}). 
Moreover, we demonstrate \method{} generalizes to an external dataset and an open-source VLM (\cref{sec:generalization}).
Lastly, we provide ablation analysis of \method{} design choices (\cref{sec:exp_ablation}).
See the appendix for more qualitative examples and ablation studies.

\begin{table}[t]
\centering
\caption{Correlation between human overall Likert scores and different metrics on \dataset{}.}
\begin{tabular}{lccc}
\toprule
Metric & Pearson's $r$ & Kendall's $\tau$ & Spearman's $\rho$\\
\midrule

VideoScore~\cite{he2024videoscore}  & 0.289 & 0.262 & 0.378 \\
VideoPhy-2-Autoeval~\cite{bansal2025videophy} & 0.346 & 0.277 & 0.349 \\
PhyGenEval~\cite{meng2024towards} & 0.272 & 0.220 & 0.264 \\
DSG~\cite{cho2024davidsonian} & 0.302 & 0.195 & 0.265 \\
Direct VQA & 0.382 & 0.290 & 0.360\\
\midrule
\method{} (Ours) \\
w/ Gemini-2.5-Pro & 0.467 & 0.306 & 0.406 \\
w/ GPT-5.5 & \textbf{0.478} & \textbf{0.336} & \textbf{0.456} \\
\bottomrule
\end{tabular}
\label{tab:correlation_human_method_comparison}
\end{table}

\subsection{Correlation to Human Judgments: Generated Video Evaluation}
\label{sec:exp_correlations}

\noindent\textbf{Correlation with Overall Human Likert Scores.}
While the focus of \method{} is to provide detailed failure mode analyses instead of providing a single scoring method, we note that its scores can be aggregated into an overall score.
Here, we experiment with \method{} as an overall video scorer for use cases where an aggregate score is desired.
We compare \method{} with recent video evaluation metrics for physical plausibility:
VideoScore~\cite{he2024videoscore},
VideoPhy-2-Autoeval~\cite{bansal2025videophy},
and PhyGenEval~\cite{meng2024towards}.
We include DSG~\cite{cho2024davidsonian}, an evaluation for image generation models, as a baseline.
For fair comparison with DSG, we use Gemini-2.5-Pro with DSG's QG/QA model components, and feed the whole video as input to the QA component, like ours.
We also include a simple direct VQA baseline; 
i.e., asking Gemini 2.5 Pro to output an alignment score between 1 and 5, given a video and a text prompt.
We evaluate Pearson's $r$, Kendall's $\tau$, and Spearman's $\rho$. 
\Cref{tab:correlation_human_method_comparison} shows that 
\method{} achieves similar or better correlation with human judgment in all three correlation coefficients.

\begin{wraptable}{r}{0.4\linewidth}
\centering
\caption{
Pearson's $r$ correlation between different human Likert score categories (per-category to overall).
}
\begin{tabular}{l ccc}
\toprule
& Object & Action & Physics \\
\midrule
Overall & 0.44 & 0.66 & 0.85 \\
\bottomrule
\end{tabular}
\label{tab:correlation_human_only_category}
\end{wraptable}

\noindent\textbf{What do human evaluators care about when measuring ``overall'' score?}
When human evaluators are asked to judge a video with a single overall score, they might not judge videos based on an unweighted average of three categories;
in other words, they may consider some categories more or less strongly in forming their final judgment (or indeed, may consider alternate factors not considered in our three categories). 
We further explore this in \Cref{tab:correlation_human_only_category}, where we study the correlation between different human Likert scores for object/action/physics categories and their overall rating for the video as a whole (see \cref{sec:dataset}).
Physics correlates most strongly with the overall score ($r = 0.85$), indicating that the annotators' final assessments are more predicted by errors in physics than other categories. 
This underscores the importance of correct physics evaluation.

\noindent\textbf{Per-category PQSG-to-human correlation.}
We study category-specific correlations between \method{}'s three category-specific scores and human's category-specific Likert scores.
For \method{}, we report two versions: one using answers from GPT, and one using human annotators.
Note that the Likert score and fine-grained question-specific evaluation have different purposes: the former is intended as an overall summary score for ranking, while the latter is designed to provide detailed feedback, with the score itself being aggregated from the answers to many granular questions.  
Because of this, we expect to see a moderate positive correlation ($>$ 0.4) instead of a very high correlation (0.9 $>$).
\Cref{tab:correlation_human_metric_category} shows this correlation in all three categories.
Among categories,
automated QA performs nearly identically to human scores in object categories, while there is some room to improve in action and physics categories. This is also consistent with our findings in QA evaluation (\Cref{tab:qa}).

\begin{table}[t]
\centering
\caption{
Pearson's $r$ correlation between different per-category human Likert scores and
per-category \method{} scores
(with both VLM and human QA).}
\begin{tabular}{l ccc}
\toprule
  & \multicolumn{3}{c}{Human Likert category - \method{} category}\\
\cmidrule(lr){2-4}
 & Object-Object & Action-Action & Physics-Physics \\
\midrule
w/ GPT-5.5 QA & 0.59 & 0.68 & 0.48 \\
w/ Human QA & 0.59 & 0.73 & 0.57\\
\bottomrule
\end{tabular}
% }
\label{tab:correlation_human_metric_category}
\end{table}

\begin{table}[t]
\centering
\begin{minipage}[t]{0.5\linewidth}
\centering
\caption{Comparison of different video generation models. \method{} scores are averaged over three question-generation runs. }
\resizebox{\linewidth}{!}{
\begin{tabular}{l ccc c}
\toprule
Video Gen. & Object & Action & Physics & Overall \\
\midrule
Sora 2     & $0.95_{\pm0.042}$ & $0.75_{\pm0.056}$ & $0.69_{\pm0.070}$ & $0.78_{\pm0.057}$ \\
Veo 3      & $0.98_{\pm0.016}$ & $0.78_{\pm0.028}$ & $0.68_{\pm0.022}$ & $0.80_{\pm0.018}$ \\
Wan 2.1    & $0.86_{\pm0.029}$ & $0.53_{\pm0.058}$ & $0.46_{\pm0.035}$ & $0.59_{\pm0.042}$ \\
Cosmos 2.5 & $0.93_{\pm0.040}$ & $0.56_{\pm0.057}$ & $0.46_{\pm0.078}$ & $0.62_{\pm0.038}$ \\
\midrule
Average & 0.93 & 0.66 & 0.57 & 0.70 \\
\bottomrule
\end{tabular}
}
\label{tab:compare_models}
\end{minipage}
\hfill
\begin{minipage}[t]{0.45\linewidth}
\centering
\caption{Human evaluation of question answering on 30 \dataset{} videos,
measured with accuracy.}
\begin{tabular}{l ccc}
\toprule
\multirow{2}{*}{QA Method} & \multicolumn{3}{c}{Accuracy (\%)} \\
\cmidrule(lr){2-4}
 & Object & Action & Physics \\
\midrule
Gemini-2.5-Pro   & 87.6 & 59.5 & 61.5 \\
GPT-5.5          & \textbf{88.4} & \textbf{63.4} & \textbf{64.6} \\
\bottomrule
\end{tabular}
\label{tab:qa}
\end{minipage}
\end{table}

\subsection{Comparing Video Generation Models via \method{}}
\label{sec:exp_comparing_video_models}

In \Cref{tab:compare_models}, we compare Sora 2, Veo 3, Wan 2.1, and Cosmos-14B on \dataset{} prompts across the three granular categories of our metric.
Here, we use automated QG and QA on the whole \dataset{}. 
Overall, we find that models struggle on action and physics, with lower scores than for object prediction.
This indicates that while models tend to generate the correct objects (some with very high accuracy), they struggle with producing the right actions and physical interactions in the video. 
Moreover, proprietary models (Sora 2 and Veo 3) outperform the open-source variants (Wan 2.1 and Cosmos-14B) by a large margin. 
The narrow confidence intervals confirm that \method{}'s model rankings remain consistent across repeated question-generation runs.

\subsection{Subtask Evaluation}
\label{sec:exp_subtasks}

As \dataset{} provides fine-grained human annotation of questions and answers, this allows the detailed evaluation of \method{} in two subtasks: question-generation (QG) and question-answering (QA).
Here, we manually annotate 20 graphs to compare VLMs to human-level graph generation, and answer generated questions to provide an upper-bound to the metric with human-level answering.

\subsubsection{Question Generation.}
\label{sec:exp_qg}
We evaluate generated \method{} questions in terms of how accurate they are (precision) and how completely they cover the different aspects of the detailed prompts used (recall).
Concretely,
given a set of human-annotated questions $Q^h={q^h_1, ..., q^h_{|Q^h|}}$,
and generated questions $Q^g={q^g_1, ..., q^g_{|Q^g|}}$,
let $m_{i} \in {0,1}$ be 1 if $q^h_i$
is semantically covered by any one or multiple generated questions from $Q^g$ 
and 0 otherwise,
and similarly, let $m_{j} \in {0,1}$ be 1 if $q^g_j$ is semantically covered by any one or multiple human annotated questions from $Q^h$ and 0 otherwise.
We manually calculate \textbf{precision} = $\sum^{|Q^g|}_{j = 1} m_{j} / |Q^g|$ and \textbf{recall} = $\sum^{|Q^h|}_{i = 1} m_{i} / |Q^h|$. 
We compare GT questions described in \cref{sec:dataset} with questions generated with two VLMs: Gemini-2.5-Pro~\cite{gemini25} and GPT-5.5~\cite{openai2025gpt5}.

\noindent\textbf{VLMs can generate highly reliable verification questions.}
As shown in \Cref{tab:qg}, \method{} questions generated by both VLMs (Gemini-2.5-Pro and GPT-5.5)
are highly aligned with human-annotated questions, in terms of
how accurate they are (precision) and how completely it covers the aspects of the detailed prompt (recall).
When examining manual annotations, the mismatches we observe are mostly due to a failure to predict future states of a prompt (such as not accounting for a possible state where objects other than the main subject of the prompt could interact). 
Even with this minor failure, generating \method{}s with VLMs is reliable with precision and recall scores above 90\%.

\begin{figure*}[t]
\centering
\includegraphics[width=\linewidth]{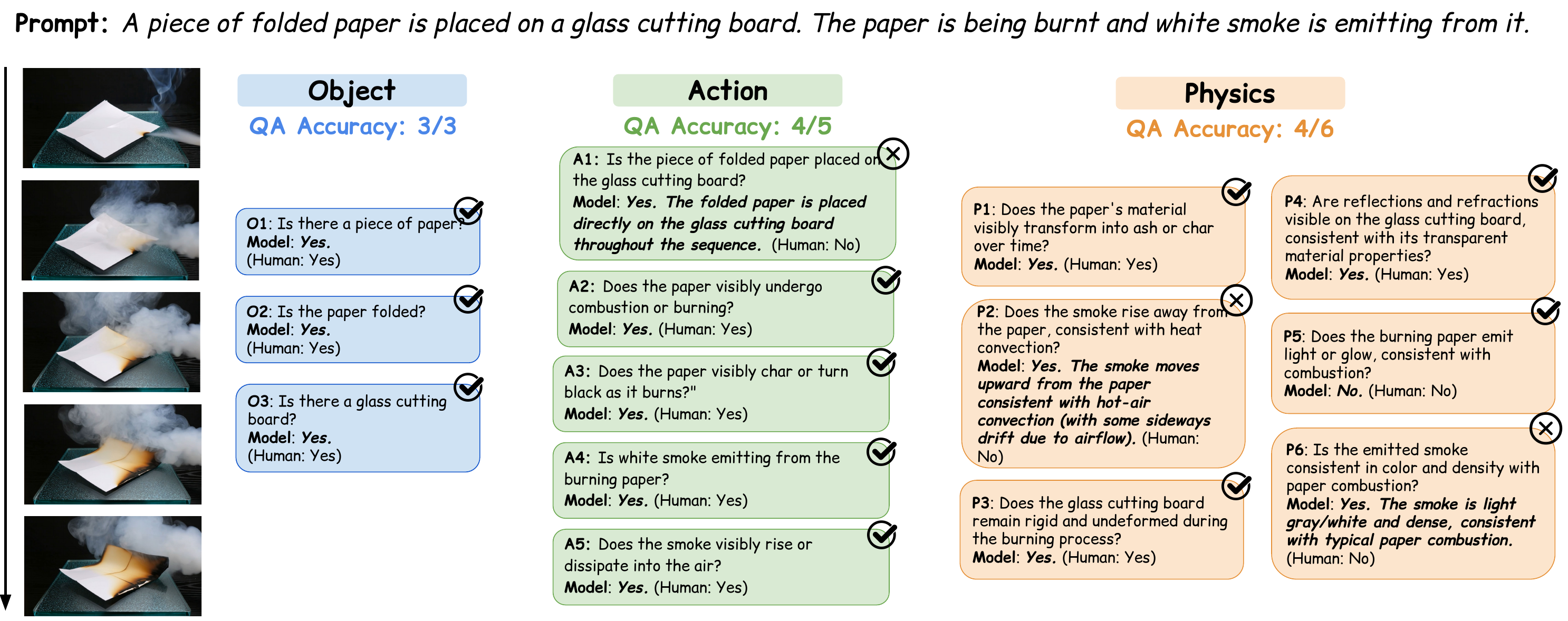}
\caption{
An example video where a QA model (GPT-5) struggles. The model fails to capture the complex dynamics of the smoke, exhibiting a strong ``yes-bias''~\cite{ross2024what, tjuatja2024llms} by defaulting answers to ``yes,'' while human answers include multiple ``no'' responses.
}
\label{fig:failure}
\end{figure*}

\begin{table}[t]
\centering
\begin{minipage}[t]{0.45\linewidth}
\centering
\caption{Human evaluation of question generation on 20 prompts.}
\begin{tabular}{lcc}
\toprule
QG Method & Precision (\%) & Recall (\%) \\
\midrule
Gemini-2.5-Pro & \textbf{95.2} & \textbf{95.2} \\
GPT-5.5 & 92.0 & \textbf{99.6}\\
\bottomrule
\end{tabular}
\label{tab:qg}
\end{minipage}
\hfill
\begin{minipage}[t]{0.45\linewidth}
\centering
\caption{
Pearson's $r$ correlation with human judgment on VideoPhy-2.
}
\begin{tabular}{l ccc}
\toprule
\textbf{Metric} & SA & PC\\
\midrule
VideoPhy-2-Autoeval & 0.450 & 0.420 \\
+ \method{} (Ours) & \textbf{0.550} & \textbf{0.498} \\
\bottomrule
\end{tabular}
\label{tab:videophy}
\end{minipage}
\end{table}

\subsubsection{Question Answering.}
\label{sec:exp_qa}

Given properly generated questions, a good question answering (QA) model is expected to answer the verification question in a way that aligns with human answers.
We examine how accurately different QA models answer each question by comparing the yes/no responses generated by the models with ground-truth answers derived from human annotations (see \cref{sec:dataset}). The comparison is evaluated using accuracy as the primary metric.
Building on the finding in \Cref{tab:qg} that models are strong question generators, we use automatically-generated questions from Gemini-2.5-Pro.
We then evaluate Gemini-2.5-Pro and GPT-5.5 as QA models. For each model, we process videos at their maximum frames-per-second (fps): 24 fps for Gemini-2.5-Pro and 8 fps for GPT-5.5.

\noindent\textbf{VLMs perform well at evaluating objects and high-level actions, but struggle with physical plausibility.}
\Cref{tab:qa} shows the accuracy of 
Gemini-2.5-Pro and GPT-5.5
models in question answering for three categories.
For object and action categories, the answer accuracy is high, but for physics, the answer accuracy is low.
This aligns with recent findings that VLMs struggle with physical reasoning \cite{chowphysbench2025} and spatio-temporal reasoning \cite{zhou2025vlm4d}. 
Manual examination of model predictions reveals common failure modes in identifying rapid actions and in capturing complex physical interactions
(e.g., the direction and density of smoke from a flame).
To further analyze the failure modes of VLMs on QA,
we provide a qualitative example in \cref{fig:failure}. 
Here, we find that the model generally suffers from ``yes-bias'' \cite{ross2024what, tjuatja2024llms}, where it tends to answer yes to questions. 
The mistakes the model (in this case, GPT-5.5) makes are also reflected in its reasoning for incorrect predictions, where it confidently claims that certain actions have occurred even though they are not present in the video.
Part of the failure on physical questions may stem from the VLM’s LLM component and the commonsense knowledge embedded in it. For example, it suggests the smoke is rising upward with a slight sideways drift due to airflow, whereas the video actually shows smoke being ejected sideways, similar to a flare. with
Taken together, these results suggest that verifying object and high-level actions in videos can be automated in high precision using recent VLMs. However, there remains significant room for improvement in evaluating low-level, rapid actions as well as assessing the physical plausibility of events in videos.

\subsection{Generalization Study: External Dataset and Open-source VLM}
\label{sec:generalization}
We demonstrate PQSG's improved performance on 100 prompt-video pairs from the VideoPhy2~\cite{bansal2025videophy} test set using the same in-context examples and prompts as other experiments.
Here, we show the generalization capability of \method{} in a different setup.
In \Cref{tab:videophy}, \method{} increases both semantic adherence (SA) and physical commonsense (PC), showing the generalization of \method{} in another dataset and with an open-source VLM.

\subsection{Iterative Refinement with \method{}}

\begin{wrapfigure}{r}{0.5\textwidth}
   \centering
\includegraphics[width=\linewidth]{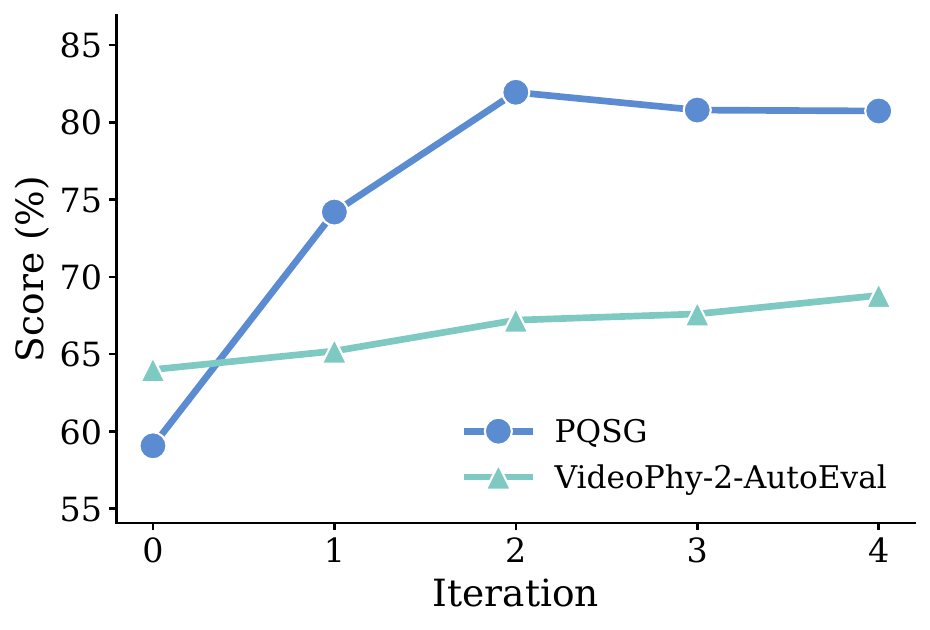}
\caption{
Scores vs. number of iterations on \method{} refinement loop. 
}
\label{fig:refinement}
\end{wrapfigure}

We investigate whether \method{} can be used to directly improve video generation.
Because \method{} decomposes quality into fine-grained questions, it enables targeted prompt refinements that explicitly emphasize missing or incorrect components in a generated video.
Following previous work in prompt refinements \cite{hao2023optimizing, manas2024improving},
we design an iterative generation loop in which an initial video is generated with Wan 2.2 TI2V-5B \cite{wan2025}
and evaluated using \method{}.
Based on the \method{} feedback, the prompt is refined by GPT-5.5 and a new video is generated.
The resulting video is then re-evaluated using the same \method{}, and the process repeats for multiple iterations.
As shown in \cref{fig:refinement}, from iteration 0 to iteration 1, the average \method{} score increases by nearly 15\%, indicating that a single round of targeted prompt refinement substantially improves generation quality. 
Performance continues to improve in iteration 2, reaching a final average score of 81.9\%, after which it plateaus with subsequent refinements.
As a baseline, we evaluate the videos Videophy2-AutoEval with the same configuration as \Cref{tab:correlation_human_method_comparison}.
The Videophy-2-AutoEval baseline shows a similar score improvement across iterations, with small gains throughout.
These results demonstrate that \method{} is not only effective as an evaluation framework, but can also be integrated into an iterative refinement loop to improve video generation quality. 
Crucially, these improvements are achieved through prompt-level augmentations guided by fine-grained \method{} feedback, without modifying the model architecture or retraining, highlighting \method{}'s utility as a practical tool for producing higher-quality videos.

\subsection{Ablation Study}
\label{sec:exp_ablation}

\begin{table}[t]
\centering
\caption{Ablation of \method{} design choices. \emph{Without fine-grained questions} represents the average of three direct VQA scores across objects, actions, and physics.}
\begin{tabular}{l ccc}
\toprule
& \multicolumn{2}{c}{Pearson's $r$} \\
\cmidrule(lr){2-3}
\textbf{Metric} & w/ GPT-5 QA & w/ Human QA\\
\midrule
\method{} (Ours) & 0.48 & 0.80 \\
Without graph dependencies & 0.44 & 0.75\\
Without fine-grained questions & 0.40 & 0.68\\
\bottomrule
\end{tabular}
\label{tab:ablation}
\end{table}

We study the design choices of \method{} by changing one component at a time, measuring Pearson's $r$ between the overall score and human Likert score.
We experiment with
adding a reference video as an additional input during the QG stage,
removing the dependency graph (i.e., not automatically marking responses to invalid question as ``no''),
and removing fine-grained questions (i.e., directly scoring 3 category scores with direct VQA and averaging).
As shown in \Cref{tab:ablation}, each of these ablations degrades performance. 
Removing the fine-grained questions and removing the dependency graph both reduce correlation with human judgment, confirming that the granular question decomposition and the logical dependency structure each contribute to PQSG's alignment with human ratings.

\section{Conclusion}
\label{sec:conclusion}

We introduced \method{}, an evaluation framework designed to address a critical gap in video generation: the lack of fine-grained, automated metrics for physical realism.
We decompose evaluation into a structured dependency graph of atomic questions, 
and collect a new dataset \dataset{}, including fine-grained human annotations of videos from strong video generation models, to facilitate research in developing reliable evaluation metrics.
In our experiments, we analyze correlations between human judgments and \method{} scores,
compare video generation methods,
verify question generation and question answering components,
and conduct ablations on design choices.

\noindent\textbf{Limitations.}
\method{}'s performance is bounded by the QA capability of the underlying VLM, where \Cref{tab:qa} demonstrates it currently achieves 64.6\% accuracy on physics questions. 
However, PQSG is model-agnostic by design: as shown in Appendix, human correlation scales with VLM capability, and when humans perform QA, Pearson's correlation reaches 0.80 (\Cref{tab:ablation}), indicating a high ceiling. 
Besides, while PQSG improves the reliability of evaluation compared to using direct VQA, 
the backbone VLM performance is still important.
We primarily use closed-source VLMs
for our experiments, which may limit reproducibility; we mitigate this by releasing all code, prompts, and annotations, and show generalization with an open-source VLM in \Cref{tab:videophy}.
Finally, PQSG only verifies interactions entailed by the prompt, so physics in regions the prompt leaves unconstrained is not assessed. Evaluating such cases without a prompt, as humans can, is a promising direction for future work.

\section*{Acknowledgments}
This work was supported by ARO Award W911NF2110220, ONR Grant N00014-23-1-2356, NSF-AI Engage Institute DRL2112635, NSF-CAREER Award 1846185, Capital One Research Award, and NVIDIA Academic Grant Program.
The views contained in this article are those of the authors and not of the funding agency.

\clearpage  % TODO FINAL: This \clearpage needs to be removed from both review and camera-ready versions.

% \section*{Acknowledgements}
% Please insert your acknowledgments here.

\bibliographystyle{splncs04}
\bibliography{main}
\appendix

\section*{Appendix}

\section{Human Annotation Details}
\label{sec:annotation_details}
In this section,  we describe our annotation framework designed to maximize objectivity, covering both the strict guidelines enforced to minimize variance and the quality control measures taken to ensure data consistency.

\subsection{Annotation Guidelines}
We provided human annotators with the strict guidelines shown in ~\cref{fig:annotation_guideline} to ensure consistent, objective, and physically grounded evaluations across all generated videos. 
Specifically, our protocol enforces three key principles:
(1) A \textbf{human-realism assumption}, instructing evaluators to judge videos against real-world standards rather than adjusting for generative model limitations; 
(2) An \textbf{absolute semantic correctness}, where annotators verify that objects and actions match the literal prompt definitions with no leeway for interpretation; and 
(3) A \textbf{decoupled physical assessment}, ensuring that physical plausibility is evaluated independently of text alignment based solely on adherence to real-world laws.

\subsection{Quality Control and Reliability}
To establish annotation reliability, we conduct a pilot study with 6 non-author undergraduate and graduate students. 
All of them were given a detailed instruction guide to complete their annotations (\cref{fig:annotation_guideline}).
Each student rated a subset of 8 or 9 videos across four dimensions (object, action, physics, and overall judgment) using 5-point Likert scales. 
In total, we collected 150 annotations across 50 videos, obtaining 3 annotations per video to compute inter-annotator agreement.

To establish the validity of our annotator task, we first measure the degree of agreement between different annotators in our study.
In \Cref{tab:agreement}, we see that annotators overall agree, with an average Intra-class Correlation Coefficient (ICC) of 0.840 and a Krippendorff's Alpha value of 0.592.
\cite{landis1977measurement} states Alpha values between 0.4 and 0.6 are ``moderate,'' and \cite{cicchetti1994guidelines} states that ICC values above 0.75 are ``excellent.'' 
Physics adherence shows the lowest agreement across all values, which is expected given the variability in judging very poor physics videos. 
Qualitatively, we find that when generated videos breaks physics laws, the annotations collected become much more opinionated and therefore noisy. 
Given that we achieve an ICC of 0.840, we conclude that \textbf{human annotations are consistent and reliable for judging the human alignment of \method{}.}

\begin{figure*}[t]
\centering
\includegraphics[width=\linewidth]{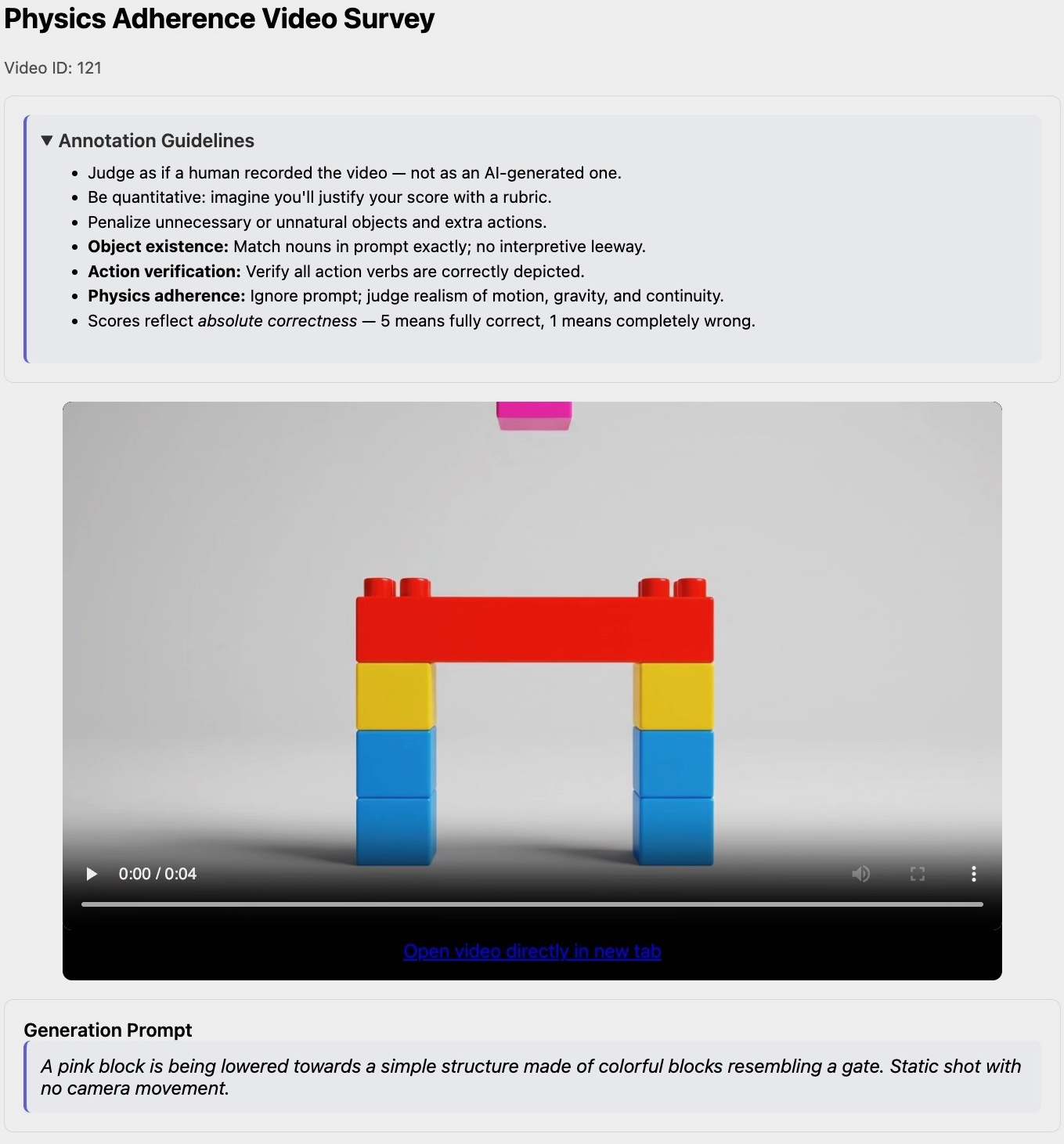}
\caption{
Partial screenshot of annotation UI given to annotators.
}
\label{fig:annotation_ui1}
\end{figure*}

\begin{figure*}[t]
\centering
\includegraphics[width=\linewidth]{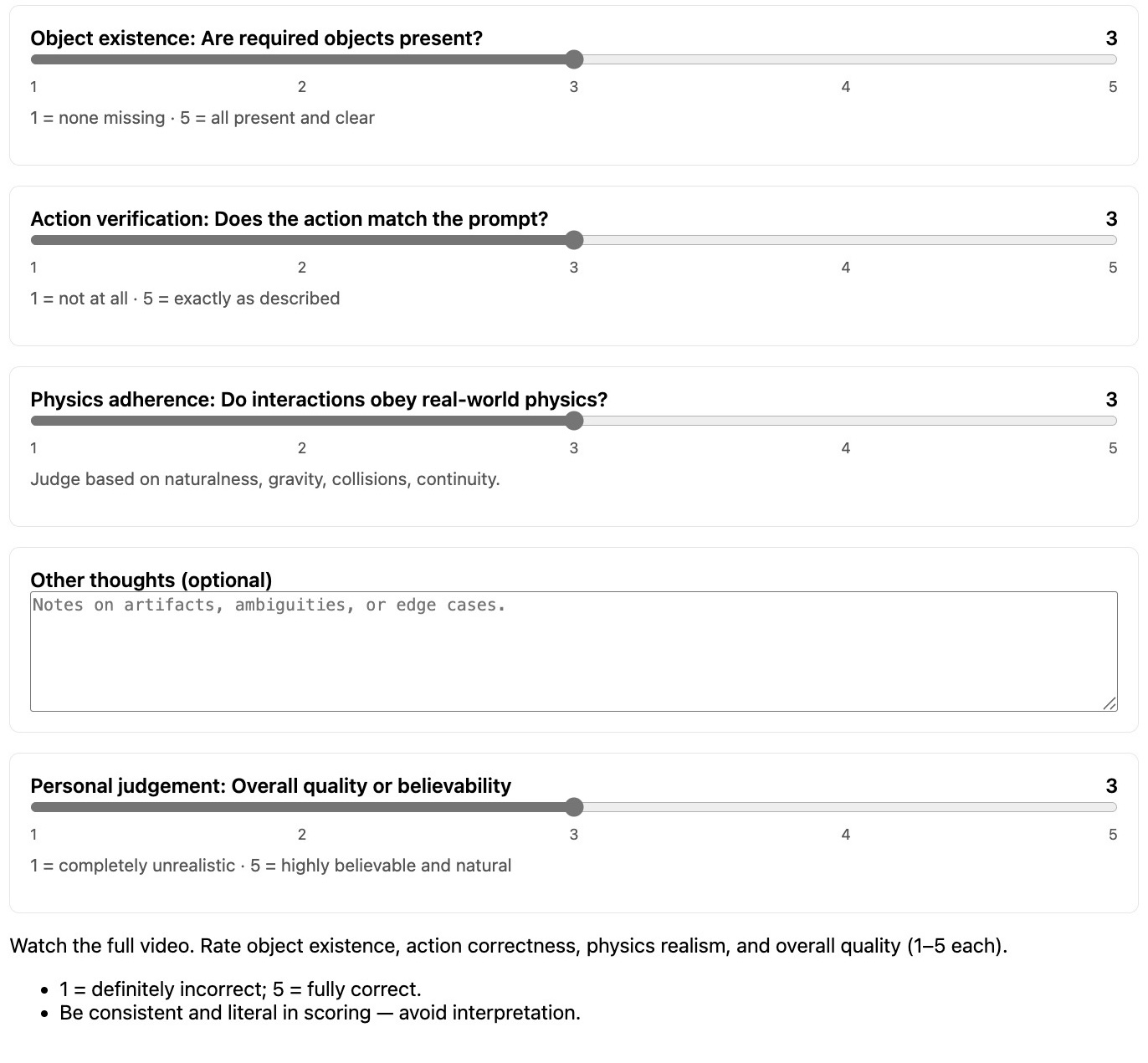}
\caption{
Partial screenshot of annotation UI given to annotators.
}
\label{fig:annotation_ui2}
\end{figure*}

After validating the annotator agreement, we collect human annotations for the remaining generated videos using the same Likert scales from a broader selection of students. 
The annotation page included guidelines at the top to remind annotators of the meanings of the different categories.
Screenshots from the annotation user interface can be viewed at \cref{fig:annotation_ui1} and \cref{fig:annotation_ui2}.

\section{Additional Experimental Results}

\subsection{QA with GPT 5.1}
In \cref{sec:exp_qa}, our findings suggest that if VLMs can improve performance in answering questions about rapid actions and intricate physical properties, we can achieve even higher correlation with humans.
Indeed, as VLMs like GPT 5.5 introduce stronger video question-answering capabilities, \Cref{tab:correlation_new_method} presents the highest Pearson's correlation of 0.478. 
These results further support the conclusion that \textbf{as the models get stronger, \method{} becomes more aligned with human judgments}.

\subsection{Ablation Study}

\begin{figure*}[t]
\centering
\includegraphics[width=\linewidth]{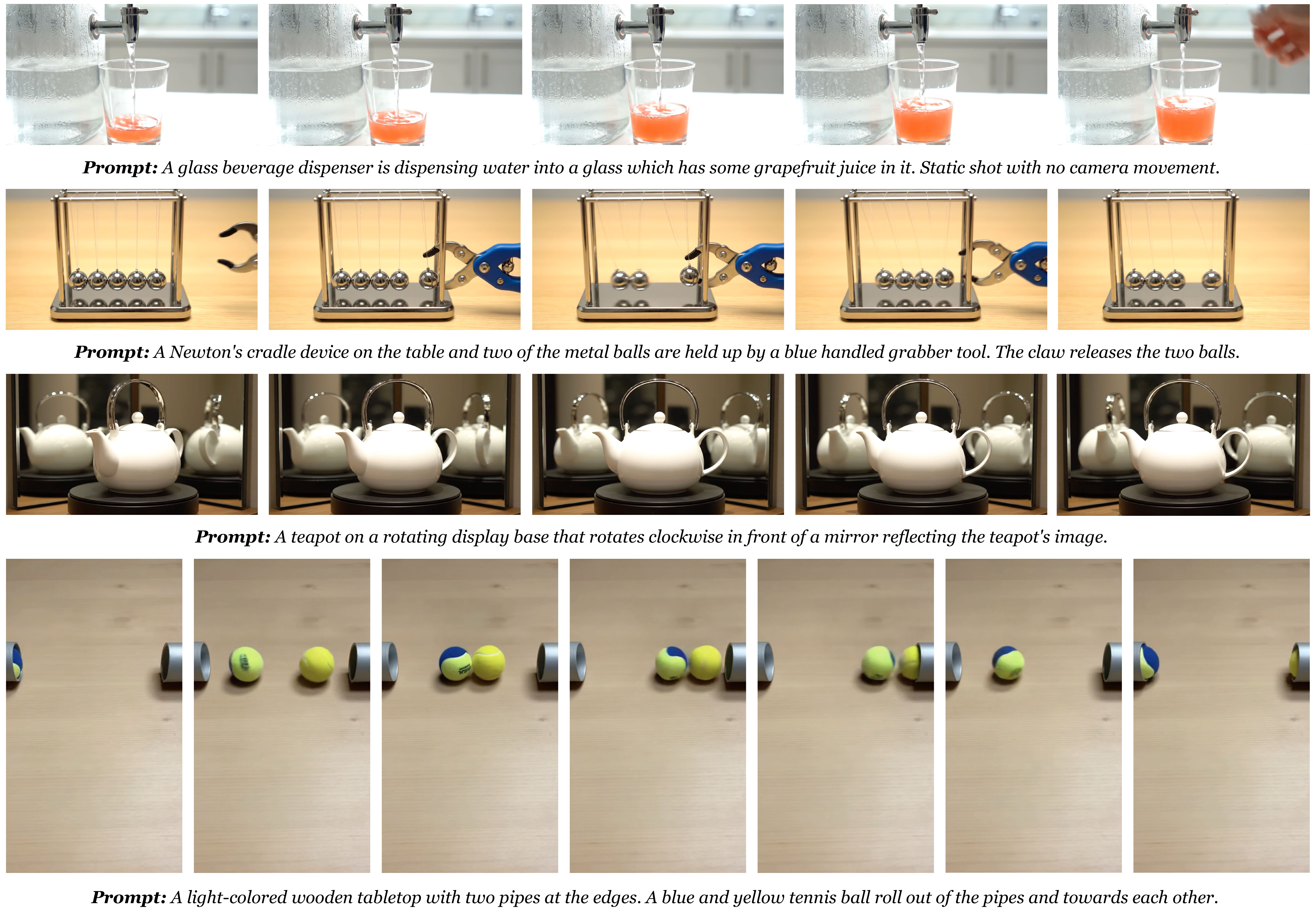}
\caption{
Example Videos from \dataset{}.
}
\label{fig:more_examples}
\end{figure*}

\begin{table}[t]
\caption{
Inter-annotator agreement on 50 videos in \dataset{}.
}
\centering
\small
\begin{tabular}{lccc}
\toprule\textbf{Dimension} & \textbf{Krippendorff's $\alpha$ ($\uparrow$)} & \textbf{ICC(2,3) ($\uparrow$)} \\
\midrule
Object & $0.588$ & $0.889$ \\
Action & $0.635$ & $0.875$ \\
Physics & $0.543$ & $0.773$ \\
Overall Judgment & $0.603$ & $0.821$ \\
\midrule
\textbf{Average} & $\mathbf{0.592}$ & $\mathbf{0.840}$ \\
\bottomrule
\end{tabular}

\label{tab:agreement}
\end{table}

\begin{table}[t]
\caption{Correlation between human overall Likert scores and different metrics on \dataset{}. Extended part of \Cref{tab:correlation_human_method_comparison} in the main paper.}
\centering
\begin{tabular}{lccc}
\toprule
\textbf{Metric} & \textbf{Pearson's} $r$ & \textbf{Kendall's} $\tau$ & \textbf{Spearman's} $\rho$\\
\midrule
w/ Gemini-2.5-Flash & 0.417 & 0.259 & 0.354 \\
w/ Gemini-2.5-Pro   & 0.467 & 0.306 & 0.406 \\
w/ GPT-5            & 0.430 & 0.263 & 0.366 \\
w/ GPT-5.4          & 0.429 & 0.274 & 0.369 \\
w/ GPT-5.5          & \textbf{0.478} & \textbf{0.336} & \textbf{0.456} \\
\bottomrule
\end{tabular}

\label{tab:correlation_new_method}
\end{table}

\begin{table}[h!]
\caption{Additional Ablation of \method{} design choices.}
\centering
\begin{tabular}{l c}
\toprule
& \multicolumn{1}{c}{Pearson's $r$} \\
\cmidrule(lr){2-2}
\textbf{Metric} & (GPT-5 QA)\\
\midrule
\method{} (Ours) & \textbf{0.478}\\
W/ Ref. Video as input to QG      & 0.418 \\
W/ Diff. Physics \& Action Prompts & 0.407 \\
\bottomrule
\end{tabular}
\label{tab:more_ablation}
\end{table}

Extending our ablation study in \cref{sec:exp_ablation},
we also experiment with two changes to the question generation phase. 
First, we attempted to add the reference video, but the model generated questions unrelated to the prompt and instead focused on irrelevant occurrences in the ground truth video. 
Second, we attempted prompt variations to further delineate the distinction between action and physics questions (e.g., by requiring action nodes to be explicit actions in the prompt and physics nodes to be inferred from the prompt), but found that the model struggled to generate coherent questions when additional prompt constraints were imposed.
We observe in \Cref{tab:more_ablation} that both of these experiments impede performance significantly.

\begin{figure*}
\begin{tcolorbox}[
    colback=gray!2,
    colframe=black!40,
    title=\textbf{Annotation Guidelines and Examples},
    fonttitle=\bfseries,
    enhanced,
    width=\textwidth,
    left=6pt,
    right=6pt,
    top=6pt,
    bottom=6pt,
]
\scriptsize

{\small\textbf{Guidelines}}\\

We provided human annotators with the following strict guidelines to ensure consistent, objective, and physically grounded evaluations across all generated videos.
\begin{itemize}[leftmargin=10pt]
    \item \textbf{Human-realism assumption:}  
    Judge videos as if they were recorded by a human. Do not adjust expectations based on generative model limitations.
    \item \textbf{Quantitative and consistent scoring:}  
    Even though detailed justification is not required per score, annotators should assume they might need to produce a rubric. Scores should follow objective and reproducible criteria.
    \item \textbf{Absolute correctness:}  
    Penalize extra actions or unnatural objects. A correct video should contain exactly what the prompt specifies—no more, no less.
    \item \textbf{Object existence:}  
    Use the most direct interpretation of object identity. Give no leeway for ambiguous shapes; score strictly based on whether nouns from the prompt appear accurately.
    \item \textbf{Action verification:}  
    Check all action verbs in the prompt and verify fulfillment.
    \item \textbf{Physics realism:}  
    Independent of the prompt. Assess whether the video follows real-world physical behavior and whether the scene is plausible in reality.
\end{itemize}

{\small\textbf{Examples}}\\

{\small\textbf{Example 1}}

\textbf{Prompt:}  
A grabber arm is holding a tennis ball above a piece of cardstock propped up on a rotating platform sitting on a table that rotates clockwise. The grabber lowers the ball and places it on the table as the cardstock rotates. Static shot with no camera movement.

\textbf{Video:} \url{Example 1 Video URL}

\begin{itemize}[leftmargin=10pt]
    \item \textbf{Object Score: 5}  
    All objects mentioned in the prompt appear correctly.
    \item \textbf{Action Score: 4}  
    All actions are satisfied, but the rotating platform moves in the wrong direction, so one point is deducted.
    \item \textbf{Physics Score: 4}  
    The ball's impact is slightly unnatural, but overall physical behavior is plausible.
\end{itemize}

{\small\textbf{Example 2}}

\textbf{Prompt:}  
A Newton's cradle device on the table and two of the metal balls are held up by a blue handled grabber tool. The claw releases the two balls. Static shot with no camera movement.

\textbf{Video:} \url{Example 2 Video URL}

\begin{itemize}[leftmargin=10pt]
    \item \textbf{Object Score: 2}  
    The Newton’s cradle is missing, and the grabber looks incorrect. Two points awarded because the metal balls are rendered correctly.
    \item \textbf{Action Score: 2}  
    The balls are not held or released; none of the expected actions occur. Two points because a Newton’s cradle device is present.
    \item \textbf{Physics Score: 3}  
    Aside from unnatural ball movements, the rest of the scene is physically plausible.
\end{itemize}

{\small\textbf{Example 3}}

\textbf{Prompt:}  
Two rows of alternating black and white dominoes are set up on a wooden table with a gap between the two rows. A wooden stick attached to a rotating platform rotates clockwise and knocks the first domino in the first row. Static shot with no camera movement.

\textbf{Video:} \url{Example 3 Video URL}

\begin{itemize}[leftmargin=10pt]
    \item \textbf{Object Score: 2}  
    Dominoes and stick exist but look unrealistic. One point for partial correctness, but other objects are incorrect.
    \item \textbf{Action Score: 1}  
    None of the described actions occur.
    \item \textbf{Physics Score: 1}  
    The scene displays highly unnatural motion and violates basic physics.
\end{itemize}

\end{tcolorbox}
\caption{Annotation Guidelines and Examples}
\label{fig:annotation_guideline}
\end{figure*}

\section{Additional \dataset{} Details and Video Examples}
\begin{table}[ht]
\centering
\caption{\dataset{} statistics.}
\resizebox{\linewidth}{!}{
\begin{tabular}{l c}
\toprule
\# Prompts & 65 (from Physics-IQ) \\
Physics category  & \multirow{2}{*}{Solid Mechanics (38), Fluid Dynamics (15), Optics (8), Thermodynamics (3), Magnetism (2)}\\
distribution per prompt  (\#) \\ 
\# Video models & 3 (Sora2 / Veo3 / Wan 2.1) \\
\# Total videos & 195 = 3 × 65 \\
\# Likert annotations per video & 4 (object, action, physics, overall) \\
\# Total Likert annotations & 780 = 4 × 195  \\
Avg. video size & 1.8MB\\
Avg. video length & 4.39s\\
Video frame rate & Sora2: 30fps, Veo3: 24fps, Wan2.1: 16fps\\
Video resolution & Sora2: 720x1080, Veo3: 1280x720, Wan2.1: 1280x720\\
\bottomrule
\end{tabular}
}
\label{tab:data_stats}
\end{table}

We provide additional examples of videos from \dataset{} in \cref{fig:more_examples} to illustrate the limited physics-rendering capabilities of video diffusion models.
In the first video, the dispenser's water level is below the spout (so the water should not flow through it), and the clear liquid turns orange in the glass. 
In the second video, the traditional Newton's cradle motion of the balls at the edge swinging back and forth is entirely misrepresented. Additionally, the model changes the number of metal balls present in various frames, making the video much more unrealistic.
In the third video, the teapot spout is not rendered correctly in the right mirror, resulting in a physically implausible reflection.
In the last video, the tennis ball collisions are erratic and random, completely violating collision laws. 
Full dataset statistics are reported in Table~\ref{tab:data_stats}.

\section{Prompts}
We include the prompts for PQSG Question Generation (QG) and Question Answering (QA) in \cref{fig:qg_prompt} and \cref{fig:qa_prompt} respectively. Additionally, we include an example input to PQSG Question Answering (QA) in \cref{fig:qg_prompt_example}.

\begin{figure*}[t]
\begin{tcolorbox}[
    colback=gray!2,
    colframe=black!40,
    title=\textbf{Prompt for PQSG Question Generation.},
    fonttitle=\bfseries,
    enhanced,
    width=\textwidth,
    left=6pt,
    right=6pt,
    top=6pt,
    bottom=6pt,
]
\scriptsize
SYSTEM / INSTRUCTIONS

You are a graph-construction assistant.

Given a short video description, output a \textbf{single valid JSON object} with exactly two top-level keys: ``nodes'' and ``edges''.

\begin{enumerate}
  \item \textbf{Nodes}

  \begin{itemize}
    \item Three subsections, in this exact order/spelling:

    ``object\_existence'', ``action\_verification'', ``physics''.

    \item Inside each subsection, list ``ID'': ``Question'' pairs.

    \begin{itemize}
      \item Object-existence IDs: O1, O2, \ldots
      \item Action-verification IDs: A1, A2, \ldots
      \item Physics IDs: P1, P2, \ldots
    \end{itemize}

    \item For each subsection create unique yes/no questions that would all be answered ``yes'' for a correct video.

    \begin{itemize}
      \item Use concise present-tense wording (``Is \ldots?'', ``Does \ldots?'').
      \item Don't include nodes related to the ``Static shot with no camera movement.'' part of the prompt.
      \item Ask questions generally, don't say ambiguous terms like ``in this video'', ``in this frame'', or ``clearly''.
    \end{itemize}
  \end{itemize}

  \item \textbf{Edges}

  \begin{itemize}
    \item Provide an array of objects with ``from'' and ``to'' keys.
    \item ``from'' is \textbf{always} an array of one or more IDs; ``to'' is a single ID.
    \item Order:
    \begin{enumerate}
      \item Object $\to$ Action links first (connect every action to its prerequisite objects).
      \item Action $\to$ Physics links next (connect physics checks to their responsible actions).
    \end{enumerate}
    \item No circular references and no other edges allowed
    \item Only add edges if the parent question is ABSOLUTELY NECESSARY to answer the child question. As in, there is ABSOLUTELY NO scenario where a video could be generated to satisfy the child if the parent were not true.
  \end{itemize}

  \item \textbf{Formatting}

  \begin{itemize}
    \item Return \textbf{only} JSON (no extra commentary).
    \item Use double quotes everywhere, no trailing commas.
    \item Two-space indentation (or minified JSON is acceptable).
  \end{itemize}
\end{enumerate}
(Examples on next page)

\end{tcolorbox}
\caption{Prompt for PQSG Question Generation. Full prompt is in the supplementary material. Examples are on next page.}
\label{fig:qg_prompt}
\end{figure*}

\begin{figure*}[t]
\begin{tcolorbox}[
    colback=gray!2,
    colframe=black!40,
    title=\textbf{Prompt for PQSG Question Generation.},
    fonttitle=\bfseries,
    enhanced,
    width=\textwidth,
    left=6pt,
    right=6pt,
    top=6pt,
    bottom=6pt,
]
\scriptsize

\textbf{EXAMPLE 1}

\textbf{Input video prompt}

``A water dispenser is dispensing water into a glass which has orange juice in it. ''

\textbf{Expected JSON output}

\begin{lstlisting}[
    breaklines=true,
    breakatwhitespace=true,
    basicstyle=\tiny\ttfamily
]
{
  "nodes": {
    "object_existence": {
      "O1": "Is there a glass beverage dispenser?",
      "O2": "Is a spout or outlet visible on the beverage dispenser?",
      "O3": "Is there a drinking glass?",
      "O4": "Does the drinking glass contain orange juice at the start?",
      "O5": "Is clear water visible inside the dispenser before pouring begins?"
    },
    "action_verification": {
      "A1": "Does the beverage dispenser release water through the spout?",
      "A2": "Does the water stream enter the drinking glass without missing it?",
      "A3": "Does the liquid level in the drinking glass rise as pouring continues?",
      "A4": "Do the incoming water and the existing orange juice visibly mix over time?",
      "A5": "Does the water stream travel downward through open air from spout to glass without interruption?",
      "A6": "Are surface ripples or small splashes visible where the water meets the juice?",
      "A7": "Does all liquid remain inside the drinking glass with no overflow?"
    },
    "physics": {
      "P1": "Does the water stream flow in a continuous column until it hits the juice?",
      "P2": "Do droplets of liquid splash upward from the juice's surface on impact?",
      "P3": "Does the mixture lighten progressively as dilution proceeds?",
      "P4": "Is the volume lost from the dispenser approximately equal to the volume gained in the glass?",
      "P5": "Do the dispenser and drinking glass maintain rigid shapes without visible deformation during pouring?",
      "P6": "Does fluid flow out of the opening if the fluid level inside the container is above the spout, creating pressure for the water to leave the spout?",
      "P7": "Are refractions and reflections on both glass surfaces sharp and consistent with transparent glass?",
      "P8": "Is the water stream transparent and refractive, distorting the background realistically?"
    }
  },
  "edges": [
    { "from": ["O1", "O2"], "to": "A1" },
    { "from": ["O3", "O5"], "to": "A2" },
    { "from": ["O3"], "to": "A3" },
    { "from": ["O4"], "to": "A4" },
    { "from": ["O4"], "to": "A6" },
    { "from": ["O3"], "to": "A7" },
    { "from": ["A2", "A5", "A6"], "to": "P2" },
    { "from": ["A4"], "to": "P3" },
    { "from": ["A3"], "to": "P4" },
    { "from": ["A1"], "to": "P6" },
    { "from": ["A5"], "to": "P8" }
  ]
}
\end{lstlisting}
....

(Additional in-context examples truncated)

....

Now produce a similar JSON for this new prompt: 
\{PROMPT\}
\end{tcolorbox}
\caption{Example in prompt for PQSG Question Generation. Full prompt is in the supplementary material.}
\label{fig:qg_prompt_example}
\end{figure*}

\begin{figure*}
\begin{tcolorbox}[
    colback=gray!2,
    colframe=black!40,
    title=\textbf{Prompt for PQSG Question Answering.},
    fonttitle=\bfseries,
    enhanced,
    width=\textwidth,
    left=6pt,
    right=6pt,
    top=6pt,
    bottom=6pt,
]
You are a verifier for AI-generated videos. Given a question and a video, it is your job to verify whether the video satisfies the following question.
\{QUESTION\}
\end{tcolorbox}
\caption{Prompt for PQSG Question Answering.}
\label{fig:qa_prompt}
\end{figure*}

\end{document}